\documentclass[10pt,twocolumn,letterpaper]{article}

\usepackage[accsupp]{axessibility}
\usepackage{iccv}
\usepackage{times}
\usepackage{epsfig}
\usepackage{graphicx}
\usepackage{amsmath}
\usepackage{amssymb}
\usepackage{booktabs}
\usepackage{multirow}
\usepackage{makecell} 
\usepackage{colortbl}
\usepackage[dvipsnames]{xcolor}

\usepackage[pagebackref=true,breaklinks=true,letterpaper=true,colorlinks,bookmarks=false]{hyperref}

\iccvfinalcopy

\ificcvfinal\fi
\definecolor{mygray}{gray}{.95}
\begin{document}

\title{~~UniTR: A Unified and Efficient Multi-Modal Transformer for ~~ Bird's-Eye-View Representation}

\author{
Haiyang Wang$^{1}$\footnotemark[1] ~~~~Hao Tang$^{1,4}$\footnotemark[1] ~~~~Shaoshuai Shi$^{2}$\footnotemark[2] ~~~~Aoxue Li$^{3}$  \\
~~~~Zhenguo Li$^{3}$  ~~~~Bernt Schiele$^{2}$  ~~~~Liwei Wang$^{1}$\footnotemark[2]   \\
{\normalsize{\hspace*{-14pt}}
}
{\normalsize
{$^1$}Peking University ~~ {$^2$}Max Planck Institute for Informatics ~~ {$^3$}Huawei, China ~~{$^4$}Pazhou Laboratory}\\
{\normalsize{\hspace*{-18pt}}
}
{\tt\small \{wanghaiyang@stu, tanghao@stu, wanglw@cis\}.pku.edu.cn}\\
{\tt\small \{sshi, schiele\}@mpi-inf.mpg.de  \{liaoxue2, Li.Zhenguo\}@huawei.com}
}

\maketitle
\renewcommand{\thefootnote}{\fnsymbol{footnote}}
\footnotetext[1]{Equal contribution.}
\footnotetext[2]{Corresponding author.}
\ificcvfinal\thispagestyle{empty}\fi

\begin{abstract}
Jointly processing information from multiple sensors is crucial to achieving accurate and robust perception for reliable autonomous driving systems. However, current 3D perception research follows a modality-specific paradigm, leading to additional computation overheads and inefficient collaboration between different sensor data. In this paper, we present an efficient multi-modal backbone for outdoor 3D perception named UniTR, which processes a variety of modalities with unified modeling and shared parameters. Unlike previous works, UniTR introduces a modality-agnostic transformer encoder to handle these view-discrepant sensor data for parallel modal-wise representation learning and automatic cross-modal interaction without additional fusion steps. More importantly, to make full use of these complementary sensor types, we present a novel multi-modal integration strategy by both considering semantic-abundant 2D perspective and geometry-aware 3D sparse neighborhood relations. UniTR is also a fundamentally task-agnostic backbone that naturally supports different 3D perception tasks. It sets a new state-of-the-art performance on the nuScenes benchmark, achieving +1.1 NDS higher for 3D object detection and +12.0 higher mIoU for BEV map segmentation with lower inference latency. Code will be available at \url{https://github.com/Haiyang-W/UniTR}.

\end{abstract}

\vspace{-5pt}
\section{Introduction}
Perceiving the physical world in 3D space is critical for reliable autonomous driving systems~\cite{bansal2018chauffeurnet, wang2019monocular}. As self-driving sensors become more advanced, integrating the complementary signals captured from different sensors (\textit{e.g.}, Cameras, LiDAR, and Radar) in a unified manner is essential. To achieve this goal, we propose UniTR, a unified yet efficient multi-modal transformer backbone that can process both 3D sparse point clouds and 2D multi-view dense images in parallel to learn the unified bird's-eye-view (BEV) representations for boosting 3D outdoor perception.

\begin{figure}[t]
\begin{center}
   \includegraphics[width=0.95\linewidth]{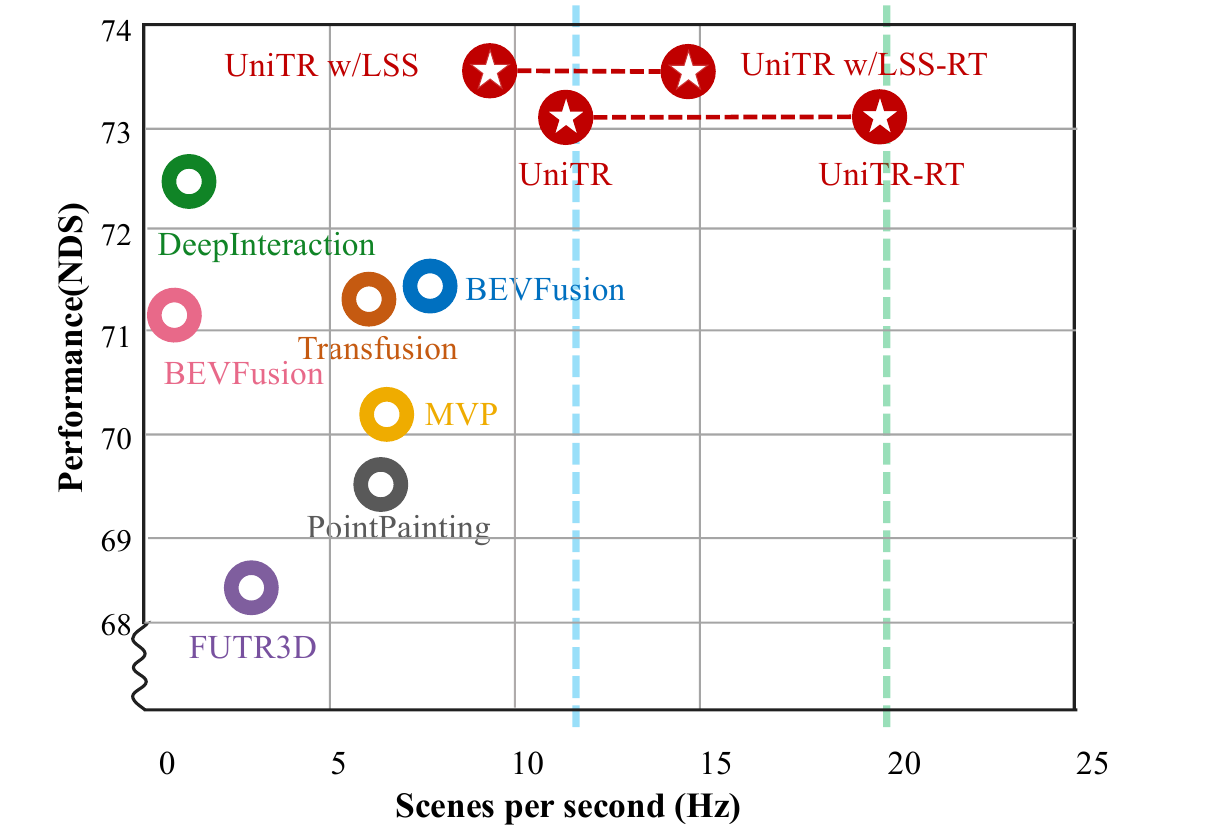}
\end{center}
\vspace{-10pt}
\caption{3D object detection performance (NDS) vs speed (Hz) on nuScenes~\cite{caesar2020nuscenes} validation set. Latency is measured on an A100 GPU with AMD EPYC 7513 CPU. {\color{SkyBlue}Blue} and {\color{Green}green} lines are the operating frequency of the camera and LiDAR in nuScenes.
}
\label{fig:speed}
\vspace{-12pt}
\end{figure}

Data obtained from multi-sensory systems are represented in fundamentally different modalities: \textit{e.g.}, cameras capture visually rich perspective images, while LiDARs acquire geometry-sensitive point clouds in 3D space.  Integrating these complementary sensors is an ideal solution for achieving robust 3D perception. However, due to the view discrepancy in raw data representations, developing an effective fusion approach is non-trivial. Previous works can be broadly classified into point-based, proposal-based, and BEV-based fusion methods. Point-based~\cite{sindagi2019mvx,vora2020pointpainting,huang2020epnet,yin2021cvpr,wang2021pointaugmenting} and proposal-based approaches~\cite{chen2017multi,yoo20203d,bai2022transfusion,li2022deepfusion,chen2022futr3d,yang2022boosting,li2023fully} enrich the LiDAR points and object proposals with semantic features from 2D images separately. BEV-based methods~\cite{liu2022bevfusion,liang2022bevfusion} unify the representations of camera and lidar into a shared BEV space and fuse them with subsequent 2D convolutions. Though performant, these fusion schemes generally require modality-specific encoders that process different sensor data in a sequential manner, leading to increased inference latency and hindering their real-world applicability. Thus, developing a modality-agnostic encoder has the potential to efficiently align the multi-modal features while facilitating the learning of generic representations for better 3D scene understanding.

Transformers have emerged as a powerful tool for multi-modal processing in various research fields~\cite{chen2020uniter,li2019visualbert,shi2022motion,zhu2022uni}.
However, its application to image-LiDAR data presents unique challenges due to the view discrepancy~(\textit{i.e.}, 2D dense images and 3D sparse point clouds). Existing transformer-based fusion strategies~\cite{bai2022transfusion,chen2022futr3d,li2022unifying} rely on modality-specific encoders followed by additional query-based late fusion, incurring non-negligible computational overheads. Hence, building an efficient and unified multi-modal backbone that can automatically learn the intra- and inter-modal representations from both image and LiDAR data is the main challenge we aim to address in this paper.
    
\begin{figure}[t]
\begin{center}
\includegraphics[width=0.95\linewidth]{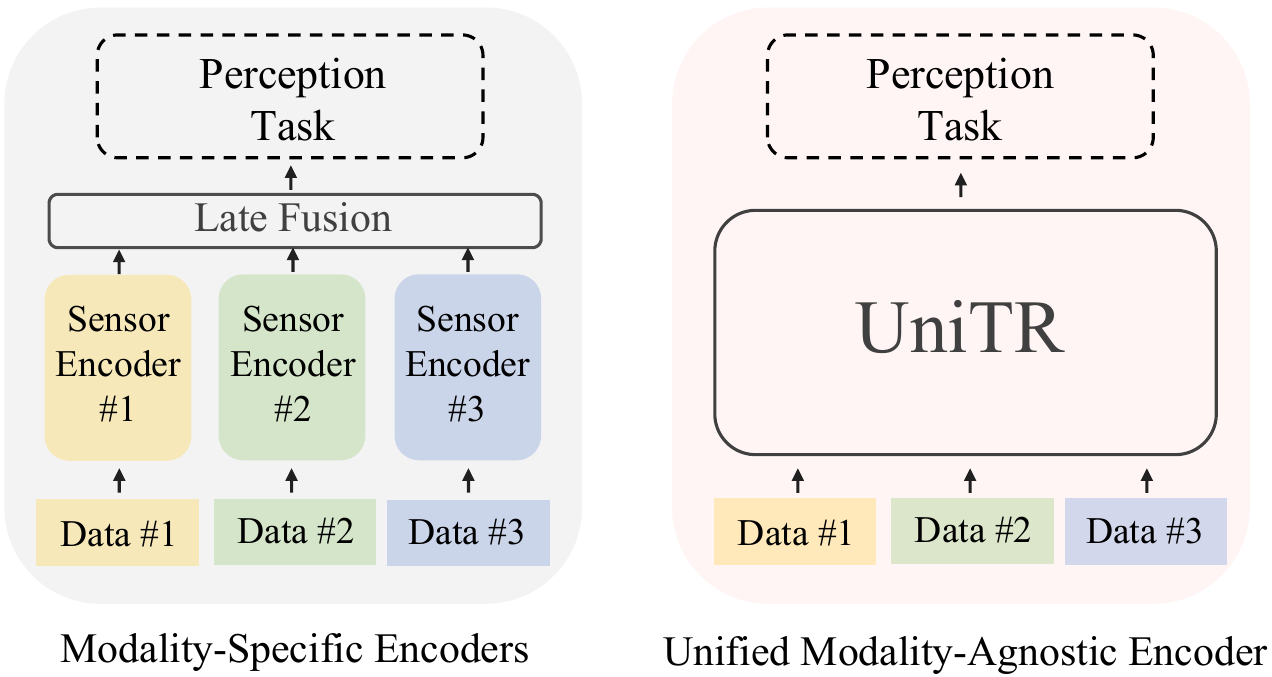}
\end{center}
\vspace{-10pt}
   \caption{Comparison between sequential modality-specific encoders and our proposed UniTR, which processes various modalities in parallel with a single model and shared parameters.}
\label{fig:compare}
\vspace{-12pt}
\end{figure}

In this paper, we introduce UniTR, a unified multi-modal transformer backbone for outdoor 3D perception. As shown in Figure \ref{fig:compare}, unlike previous modality-specific encoders, our UniTR processes the data from multi-sensors in parallel with a modal-shared transformer encoder and integrates them automatically without additional fusion steps. To achieve these goals, we design two major transformer blocks extended on DSVT~\cite{wang2023dsvt}~(\textit{i.e.}, a powerful yet flexible transformer architecture for sparse data). One is the intra-modal block that facilitates parallel computation of modal-wise representation learning for the data from each sensor, and the other one is an inter-modal block to perform cross-modal feature interaction by considering both 2D perspective and 3D geometric neighborhood relations. 

Specifically, given single- or multi-view images and point clouds,  we first convert them into unified token sequences with lightweight modality-specific tokenizers, \textit{i.e.}, 2D Convolution~\cite{krizhevsky2017imagenet} for images and voxel feature encoding layer~\cite{zhou2020end} for point clouds. To jointly process the modal-wise representation learning of each sensor, these sequences are then partitioned to size-equivalent local sets in their corresponding modal space separately, which are then assigned to different samples in the same batch for parallel computation by several modality-agnostic DSVT blocks. This strategy maximizes the parallel computing capabilities of modern GPU, which greatly reduces the inference latency~(about 2$\times$ faster) 
 while also achieving better performance by sharing weights among different modalities~(see Table \ref{tab:singlemodal}). 

Secondly, we present a powerful yet efficient cross-modal transformer block for camera-lidar fusion. To resolve the view discrepancy, previous methods adopt two view transformations, \textit{i.e.}, LiDAR-to-camera~\cite{bai2022transfusion,yang2022boosting,chen2022futr3d,li2022deepfusion} or camera-to-LiDAR projection~\cite{liu2022bevfusion,liang2022bevfusion,philion2020lift}, to unify multi-modal features in a shared space. However, these two fusion spaces are actually complementary due to their distinct neighborhood relations among the inputs, \emph{i.e.}, 2D camera view preserves dense semantic relations, while 3D lidar space provides sparse geometric structures. Combining them can benefit the performance of both geometric- and semantic-oriented 3D perception. More importantly, the time cost brought by additional late fusion behind the modality-specific encoders is normally considerable. To address these problems, we design an inter-modal transformer block to bridge different modalities according to 2D and 3D structural relations. Equipped with this block, our UniTR can integrate multi-modal features with alternative 2D-3D neighborhood configurations during backbone processing, and the performance gains (see Table \ref{tab:fuse}) demonstrate its effectiveness. Notably, this block is also built upon DSVT and can be seamlessly inserted into our multi-modal backbone.

In a nutshell, our contribution is four-fold. 1) We present a weight-sharing intra-modal transformer block for efficient modal-wise representation learning in parallel.  2) To bridge sensors with disparate views, a powerful cross-modal transformer block is designed for integrating different modalities by considering both 2D perspective and 3D geometric structural relations.  3) With the above designs,  we introduce a novel multi-modal backbone for outdoor 3D perception named UniTR, which processes a variety of modalities with shared parameters in a unified manner. 4) Our UniTR achieves state-of-the-art performance on nuScenes~\cite{caesar2020nuscenes} benchmark of various 3D perception tasks, \emph{i.e.}, 3D Object Detection~(+1.1) and BEV Map Segmentation~(+12.0), with lower latency, as shown in Figure \ref{fig:speed}.

We hope the observed strong performance of UniTR can serve as an encouraging benchmark for future efforts toward integrating visual and geometric signals with unified architectures for more generic outdoor 3D perception.

\section{Related Work}
\noindent \textbf{LiDAR-Based 3D Perception.} LiDAR sensors have become indispensable devices providing abundant geometric information for reliable autonomous driving. Existing lidar-based 3D perception research can be classified into two lines in terms of representations, \emph{i.e.}, point-based and voxel-based methods. Point-based approaches~\cite{shi2019pointrcnn,qi2019deep,wang2022rbgnet,yang20203dssd,cheng2021back} adopt PointNet~\cite{qi2017pointnet} and its variants~\cite{qi2018pointnnetplus,liu2019relation} to extract features from point clouds. Voxel-based methods~\cite{yan2018second,shi2020pv,yin2021cvpr,Deng2021VoxelRT,shi2021pv,shi2020p2,wang2022cagroupd,zhu2021cylindrical} first convert point clouds into regular 3D voxels and then process them with sparse convolution~\cite{graham20183d,choy20194d} or sparse voxel transformer~\cite{wang2023dsvt,liu2023flatformer,fan2021embracing,mao2021voxel,dong2022mssvt,yang2022unified,voxelset}.

\noindent \textbf{Camera-Based 3D Perception.} Perceiving 3D objects only with cameras has been heavily investigated due to the high cost of LiDAR sensors. Previous methods~\cite{wang2021fcos3d,wang2022detr3d,liu2022petr,philion2020lift,li2022bevformer} extract 3D information from camera data by adding extra 3D regression branches to 2D detectors~\cite{wang2021fcos3d}, designing DETR-based heads with learnable 3D object queries~\cite{wang2022detr3d,liu2022petr} or converting image features from perspective view to bird's-eye view~(BEV) using view transformers~\cite{philion2020lift,huang2021bevdet,li2022bevformer}. However, camera-based methods often face challenges such as limited depth information and occlusion, which require sophisticated modeling and post-processing.

\noindent \textbf{Multi-Sensor 3D Perception.} LiDAR and camera are complementary signals for achieving reliable autonomous driving, which has been well-explored recently. Previous multi-sensor 3D perception approaches can be coarsely classified into three types by fusion, \emph{i.e.}, point-, proposal- and BEV-based methods. Point-based~\cite{sindagi2019mvx,vora2020pointpainting,huang2020epnet,yin2021cvpr,wang2021pointaugmenting,wu2023virtual} and proposal-based approaches\cite{chen2017multi,yoo20203d,bai2022transfusion,li2022deepfusion,yang2022boosting,wang2023mvcontrast,lu2023cross} typically leverage image features to augment LiDAR points or 3D object proposals. BEV-based methods~\cite{liu2022bevfusion,liang2022bevfusion,gao2023sparse} efficiently unify the representations of camera and lidar into BEV space, and fuse them with 2D convolution for various 3D perception tasks. However, these successors often rely on sequential processing with modality-specific encoders, followed by additional late fusion steps,  which slows down the inference speed and limits real-world applications. To address these challenges, we present a unified and weight-sharing backbone that can efficiently learn intra- and inter-modal representations in a fully parallel manner, enabling faster inference and broader real-world applicability.

\begin{figure*}[t]
\begin{center}
\vspace{-10pt}
\includegraphics[width=0.95\linewidth]{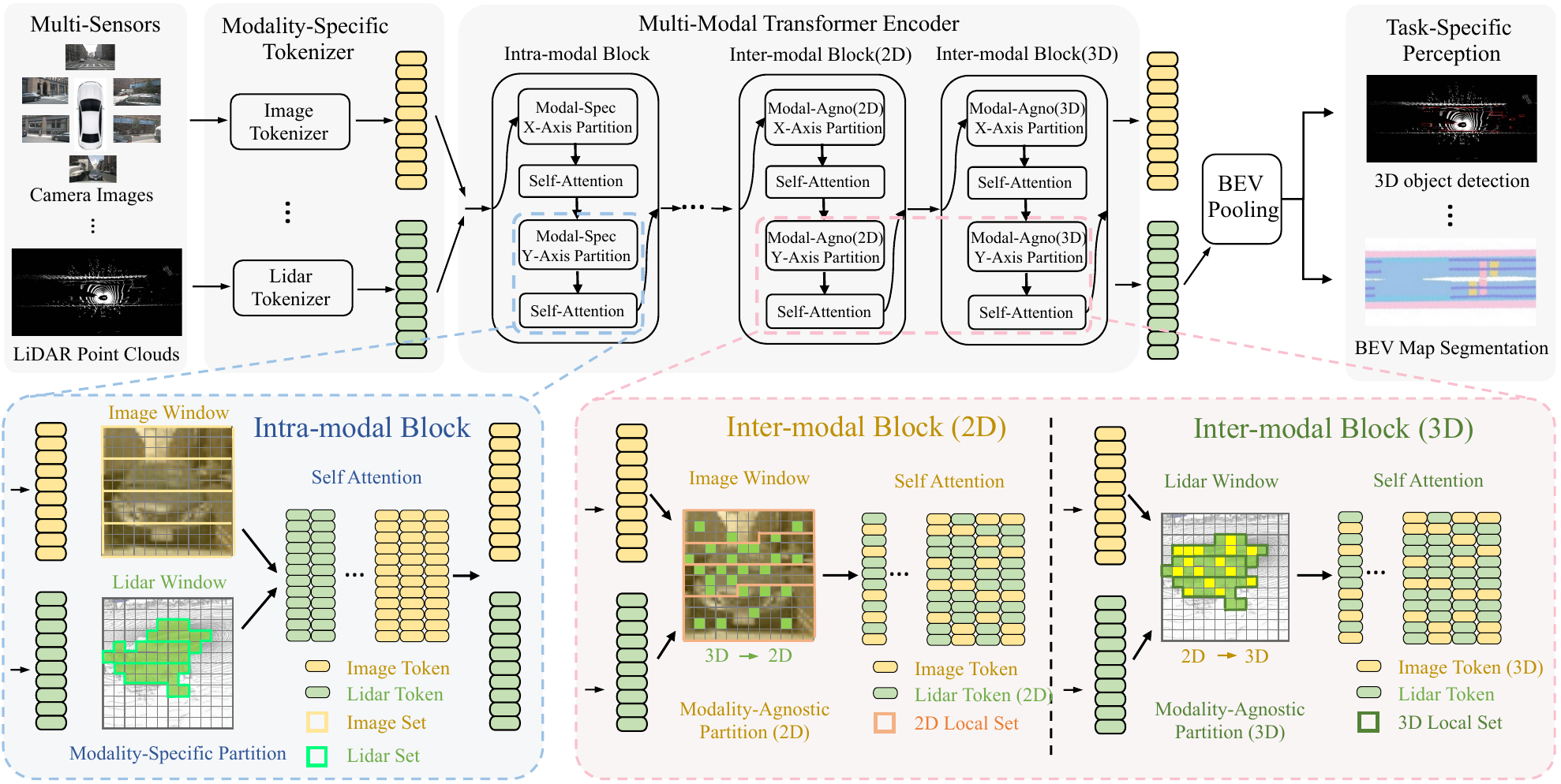}
\end{center}
   \vspace{-12pt}
   \caption{An illustration of our UniTR. Given different sensory inputs, the model first converts them into token sequences with modality-specific tokenizers. A multi-modal transformer encoder is then employed to perform single- and cross-modal representation learning and efficiently pools the semantically enriched lidar tokens into a BEV space for boosting various 3D perception tasks.}
\label{fig:overview}
\vspace{-12pt}
\end{figure*}

\section{Revisiting DSVT} \label{sec:revisit}
Dynamic Sparse Voxel Transformer~(DSVT)~\cite{wang2023dsvt} is a window-based voxel transformer backbone for outdoor 3D perception from point clouds. In order to efficiently handle sparse data in a fully parallel manner, they reformulate the standard window attention~\cite{liu2021swin} as parallel computing self-attention within a series of window-bounded and size-equivalent subsets. To allow the cross-set connection, they design a rotated set partition strategy that alternates between two partition configurations in consecutive attention layers. \\
\textbf{Dynamic set partition.} Given the sparse tokens and window partition, they further divide a series of local regions in each window based on their sparsity. As for a specific window, it has $T$ tokens, $\mathcal{T} = \{ t_i\}^T_{i=1}$. $S$ is the required number of sub-sets, which dynamically varies with the window sparsity. Then they evenly distribute $T$ tokens into $S$ sets, (\textit{e.g.}, $\mathcal{Q}_j = \{q^j_k\}^{\tau-1}_{k=0}$ is the token indices of $j$-th set). The whole process can be formulated as,
\vspace{-2pt}
\begin{equation}
  \{\mathcal{Q}_j\}_{j=0}^{S-1} = \text{DSP}(\mathcal{T}, L \times W \times H, \tau),  
  \label{eq:revisit_dsp}
  \vspace{-2pt}
\end{equation}
where $L \times W \times H$ is the window shape and $\tau$ is the size of each set, regardless of $T$, which enables the attention computation can be efficiently performed on all sets in parallel. 

After obtaining the partition $\mathcal{Q}_j$ of $j$-th set, corresponding token features and coordinates will be collected based on the predefined token inner-window id $\mathcal{I} = \{I_i\}_{i=1}^T$ as, 
\vspace{-4pt}
\begin{equation}
  \mathcal{F}_j, \mathcal{C}_j = \text{INDEX}(\mathcal{T}, \mathcal{Q}_j, \mathcal{I}),  
  \label{eq:index}
  \vspace{-4pt}
\end{equation}
where $\text{INDEX}(\cdot _{\text{voxels}}, \cdot _{\text{partition}}, \cdot _{\text{ID}})$ is the index operation, $\mathcal{F}_j \in \mathbb{R}^{\tau \times C}$ and $\mathcal{C}_j \in \mathbb{R}^{\tau \times 3} $ are the respective token features and spatial coordinates  $(x, y, z)$ of this set. In this way, they obtain some non-overlapped and size-equivalent subsets for the following parallel attention computation.  

\noindent \textbf{Rotated set attention.} 
To bridge the tokens across these non-overlapping sets, DSVT proposes the rotated-set attention that alternates between two spatially rotated partitioning configurations in successive attention layers as
\vspace{-4pt}
\begin{equation}
    \begin{aligned}
    &\mathcal{F}^{l}, \mathcal{C}^{l} = \text{INDEX}(\mathcal{T}^{l-1}, \{\mathcal{Q}_j\}_{j=0}^{S-1}, \mathcal{I}_x), \\
    &\mathcal{T}^{l} = \text{MHSA}(\mathcal{F}^{l}, \text{PE}(\mathcal{C}^{l})), \\
    &\mathcal{F}^{l+1}, \mathcal{C}^{l+1} = \text{INDEX}(\mathcal{T}^{l}, \{\mathcal{Q}_j\}_{j=0}^{S-1}, \mathcal{I}_y), \\
    &\mathcal{T}^{l+1} = \text{MHSA}(\mathcal{F}^{l+1}, \text{PE}(\mathcal{C}^{l+1})), 
    \end{aligned}
\vspace{-4pt}
\end{equation}
where $\mathcal{I}_x$ and $\mathcal{I}_y$ are the inner-window voxel index sorted in X- and Y-Axis main order respectively. $\text{MHSA}(\cdot)$ and $\text{PE}(\cdot)$ denote the Multi-head Self-Attention layer and positional encoding. $\mathcal{F} \in \mathbb{R}^{S \times \tau \times C}$ and $\mathcal{C} \in \mathbb{R}^{S \times \tau \times 3}$ are the indexed voxel features and coordinates of all sets. In this way, the original sparse window attention will be approximatively reformulated as several set attention, which can be processed in the same batch in parallel. Notably, the set partition configuration used in DSVT is generalized and flexible to be adapted to different data structures and modalities.

\section{Methodology}
In this section, we will describe our unified architecture for various modalities (\textit{i.e.}, multi-view cameras and LiDAR) and tasks (\textit{i.e.}, detection and segmentation). Figure~\ref{fig:overview} illustrates the architecture. Given different sensory inputs, the model first converts them into token sequences with modality-specific tokenizers. A modality-agnostic Transformer backbone is then employed to perform single- and cross-modal representation learning~(\S\ref{sec:single-modal}-\S\ref{sec:cross-modal}) for boosting various 3D perception tasks~(\S\ref{sec:multi-task}).
\subsection{Single-Modal Representation Learning} \label{sec:single-modal}
Perceiving 3D scenes in autonomous driving scenarios requires reliable representations of multiple modalities, \emph{e.g.}, multi-view images and sparse point clouds. Due to their discrepant representations, previous approaches~\cite{li2022unifying,chen2022futr3d,liu2022bevfusion,liang2022bevfusion} encode features of each modality by separate encoders that are generally processed sequentially, slowing down the inference speed and limiting their real-world applications. To tackle these problems, we propose to process intra-modal representation learning of each sensor with a unified architecture in parallel, whose parameters are shared for all modalities. Our approach begins by converting different modal inputs into token sequences using modality-specific tokenizers, followed by several modal-shared DSVT blocks that enable parallel modal-wise feature encoding. 

\noindent \textbf{Tokenization.} Given the raw images $X^I$ captured from $\mathcal{B}$ cameras and point clouds $X^P$ obtained from LiDAR, modality-specific tokenizers are applied to generate the input token sequences for the succeeding transformer encoder. Specifically, we use the image patch tokenizer~\cite{dosovitskiy2020image} for image modality and the dynamic voxel feature encoding tokenizer~\cite{zhou2020end} for point cloud modality. As illustrated in Figure \ref{fig:overview}, by adopting these two tokenizers, the input sequence $\mathcal{T} \in \mathbb{R}^{(M+N) \times C}$ of the following intra-modal transformer block can be composed of $N$ point cloud tokens $\mathcal{T}^P \in \mathbb{R}^{N \times C}$ and $M$ image tokens $\mathcal{T}^I \in \mathbb{R}^{M \times C}$. More details of these tokenizers are described in Appendix. 

\noindent \textbf{Modality-specific set attention.} To efficiently process intra-modal representation learning among the given multi-modal tokens in parallel, we first perform the dynamic set partition introduced in \cite{wang2023dsvt} for each sensor individually.

Specifically, after obtaining the lidar and image tokens from a specific scene, \textit{i.e.},
\vspace{-3pt}
\begin{equation}
  \begin{aligned}
  & \mathcal{T}^P  = \{ t_i^P | t_i^P= [(x_i^P, y_i^P, z_i^P); f_i^P] \}^N_{i=1}, \\
  & \mathcal{T}^I  = \{ t_i^I | t_i^I= [(x_i^I, y_i^I, b_i^I); f_i^I] \}^M_{i=1},
  \end{aligned}
  \vspace{-3pt}
\end{equation}
where $(x_i^{(*)}, y_i^{(*)}, z_i^{(*)}) \in \mathbb{R}^3$ and $f_i^{(*)} \in \mathbb{R}^C$ denote the coordinates and features of lidar and image tokens. $b_i^I \in [0, \mathcal{B}-1]$  is the corresponding view ID of the $i$-th image patch. We compute the token indices of each local set in its respective modal space as follows,
\vspace{-3pt}
\begin{equation}
  \begin{aligned}
  & \{\mathcal{Q}^P_n\}_{n=0}^{\mathcal{N}}  = \text{DSP}(\mathcal{T}^P, L^P \times W^P \times H^P, \tau), \\
  & \{\mathcal{Q}^I_m\}_{m=0}^{\mathcal{M}}  = \text{DSP}(\mathcal{T}^I, L^I \times W^I \times 1, \tau), 
  \end{aligned}
  \vspace{-3pt}
\end{equation}
where $\text{DSP}(\cdot _{\text{input}}, \cdot _{\text{window size}}, \cdot _{\text{token number of each set}})$ is the standard dynamic set partitioning strategy introduced in DSVT~\cite{wang2023dsvt}. $(L^{(*)}, W^{(*)}, H^{(*)})$ and $\tau$ are the window size and the token number of each set used in this step. $\mathcal{N}$ and $\mathcal{M}$ are the number of partitioned subsets for lidar and image respectively. Note that image window size $(L^I \times W^I \times 1)$ indicates that the image set partition is only performed within each camera view individually.

Given the modal-wise image-lidar partition, $\{\mathcal{Q}^P_n\}_{n=0}^{\mathcal{N}}$ and $\{\mathcal{Q}^I_m\}_{m=0}^{\mathcal{M}}$, we then collect the token features and coordinates assigned to each set and perform parallel attention computation for each modality as,
\vspace{-3pt}
\begin{equation}
    \begin{aligned}
    &\mathcal{F}_{l}, \mathcal{C}_{l} = \text{INDEX}([\mathcal{T}^P_{l-1}, \mathcal{T}^I_{l-1}], [\{\mathcal{Q}^P_n\}_{n\text{=}0}^{\mathcal{N}}, \{\mathcal{Q}^I_m\}_{m\text{=}0}^{\mathcal{M}}]), \\
    &~~~~~~~~~~~~~~\widetilde{\mathcal{T}}^P_{l}, \widetilde{\mathcal{T}}^I_{l} = \text{MHSA}(\mathcal{F}_{l}, \text{PE}(\mathcal{C}_{l})), \\
    \end{aligned}
    \vspace{-5pt}
\end{equation}
where $\text{INDEX}$ is the index function described in Eq.~\eqref{eq:index}~(we remove the inner-window ID here for simplicity) and $[\cdot, \cdot]$ means concatenation operation. $\mathcal{F} \in \mathbb{R}^{(\mathcal{M} + \mathcal{N}) \times \tau \times C}$ and $\mathcal{C} \in \mathbb{R}^{(\mathcal{M} + \mathcal{N}) \times \tau \times 3} $ are the multi-modal token features and spatial coordinates. $\text{MHSA}(\cdot)$ denotes the standard Multi-head Self-Attention layer with FFN and Layer Normalization, propagating information among $\tau$ tokens in each local set. $\text{PE}(\cdot)$ stands for the positional encoding function. In this way, the modal-wise representation learning will be performed by a series of modality-specific set attention in parallel. Notably, our model shares self-attention weights across modalities, enabling parallel computation and making it more computationally efficient (about $2 \times$ faster) than processing each modality separately~(see Table \ref{tab:singlemodal}).

\subsection{Cross-Modal Representation Learning} \label{sec:cross-modal}
To effectively integrate the view-discrepant information from multiple sensors in autonomous driving scenarios, existing methods typically involve designing separate deep models for each sensor and fusing the information via post-processing approaches, such as augmenting 3D object proposals~\cite{bai2022transfusion} or lidar-only BEV maps~\cite{liu2022bevfusion,liang2022bevfusion} with semantic features from image views. Moreover, prior to the fusion step, all the sensor data are usually converted into a unified representation space~(\emph{i.e.}, 3D/BEV space~\cite{liu2022bevfusion,liang2022bevfusion} or 2D perspective space~\cite{bai2022transfusion,yang2022boosting}), which are either semantic-lossy or geometry-diminished. To allow the efficient cross-modal connection and make full use of these two complementary representations, we design two partitioning configurations alternated in consecutive modality-agnostic DSVT blocks, which can automatically fuse the multi-modal data by considering both the 2D and 3D structural relations.

\noindent \textbf{Image perspective space.} To bridge the multi-sensor data in semantic-aware 2D image space, we first transform all the lidar tokens to the image plane by utilizing the camera's intrinsic and extrinsic parameters. To create a one-to-one projection for the succeeding unified perspective partition, we only take the first view hit of 3D lidar tokens and place them in their respective 2D positions as follows,
\vspace{-2pt}
\begin{equation}
\mathcal{T}^P \rightarrow \mathcal{T}^P_{2D}: (x^P, y^P, z^P) \rightarrow (x_{2D}^{P}, y_{2D}^{P}, b_{2D}^{P}),
\vspace{-2pt}
\end{equation}
where $b_{2D}^{P}$ is the identified view ID of each lidar token.

After unifying both the image and lidar tokens into camera perspective space, we use the conventional dynamic set partition to generate $\widetilde{\mathcal{M}}$ cross-modal 2D local sets in a modality-agnostic manner, \textit{i.e.},
\vspace{-2pt}
\begin{equation}
\{\mathcal{Q}^{2D}_m\}_{m=0}^{\widetilde{\mathcal{M}}}  = \text{DSP}([\mathcal{T}^P_{2D}, \mathcal{T}^I], L^I \times W^I \times 1, \tau),
\vspace{-2pt}
\end{equation}
where $[\cdot, \cdot]$ denotes concatenation. This step employs a modality-agnostic manner to group multi-modal tokens into the same local sets based on their positions in camera perspective space. Then these modality-mixed subsets are used by several DSVT blocks for 2D cross-modal interaction.

\noindent \textbf{3D geometric space.} To unify the multi-modal inputs in 3D space, an efficient and robust view projection is required to uniquely map the image patches into 3D space. However, depth-degraded camera-to-lidar transformation is an ill-posed problem due to the inherent ambiguity of the depth associated with each image pixel. Although previous learnable depth estimators~\cite{godard2017unsupervised,fu2018deep} can predict a depth image with acceptable accuracy, they require an additional computation-intensive prediction module and suffer from poor generalization ability. To overcome these limitations, inspired by MVP~\cite{yin2021multimodal}, we present a non-learnable and totally pre-computable approach to efficiently transform image patches into 3D space for the succeeding partition. 

Specifically, we first sample a group of pseudo 3D grid points, $\mathcal{V}^{P} \in \mathbb{R}^{L^S \times W^S \times H^S \times 3}$, where $L^S, W^S, H^S$ are the spatial shape of the 3D space divided by predefined grid size. Then we project all the pseudo lidar points, $\mathcal{V}^{P}$, into their corresponding virtual image coordinates (denote as $\mathcal{V}^{I}=\{v_k | (x_k, y_k); b_k; d_k)\}_{k=0}^{\nu}$), where $b_k$ and $d_k$ are the view ID and associated depth. Notably, only the projected points that fall within the view images will be considered, so the number of valid image points $\nu \le |\mathcal{V}^P| \times \mathcal{B}$.

With these pseudo image points $\mathcal{V}^I$, we retrieve the depth estimate of each image token from its nearest image virtual neighbor as follows,
\vspace{-2pt}
\begin{equation}
\{v^{\text{nearest}}_i|(x_i, y_i);b_i;d_i\}_{i=0}^M  = \text{Nearest}( \mathcal{V}^I, \mathcal{T}^{I}),
\vspace{-2pt}
\end{equation}
where $\text{Nearest}(\cdot_{\text{virtual points}}, \cdot_{\text{image tokens}})$ is the nearest neighbor function and $d_i$ represents the computed depth of $i$-th image token, which is the same as its nearest neighbor. Then we unproject the image tokens $\mathcal{T}^{I}$ back to 3D space and generate offset features based on their 2D distance, \emph{i.e.}, 
\vspace{-2pt}
\begin{equation}
\begin{aligned}
& \mathcal{T}^I \rightarrow \mathcal{T}^I_{3D}: (x^I, y^I, b^I) \rightarrow (x_{3D}^{I}, y_{3D}^{I}, z_{3D}^{I}), \\
& f^{I} \rightarrow f^{I} + \text{MLP}(||v^{\text{nearest}}(x, y) - (x^I, y^I)||).
\end{aligned}
\vspace{-2pt}
\end{equation}
In this way, we unify the image and lidar tokens in 3D space by a pre-calculable view projection, which can be totally cached during inference. Finally, a dynamic set partition module is applied to generate 3D cross-modal local sets as 
\vspace{-2pt}
\begin{equation}
\{\mathcal{Q}^{3D}_n\}_{n\text{=}0}^{\widetilde{\mathcal{N}}}  = \text{DSP}([\mathcal{T}^P, \mathcal{T}^I_{3D}], L^P\times W^P\times H^P,\tau),
\vspace{-2pt}
\end{equation}
which is then processed by a DSVT block in a unified manner for cross-modal interaction in 3D lidar space.

Our cross-modal transformer block introduces connections between different modalities by considering both 2D and 3D structural relations and is found to be effective in multi-modal 3D perception, as shown in Table \ref{tab:fuse}. Importantly, it shares the same fundation~(DSVT) of intra-modal block and can be integrated into our backbone seamlessly.

\begin{table*}[h]
\begin{center}
\resizebox{0.9\linewidth}{!}{
\begin{tabular}{l|c|c|cccc |>{\columncolor{mygray}}c}
\toprule
Methods & Present at&  Modality & NDS (\textit{val}) & mAP (\textit{val}) & NDS (\textit{test}) & mAP (\textit{test}) & Latency (\textit{ms}) \\
\midrule
BEVFormer~\cite{li2022bevformer} & ECCV'22 & C & - & -  & 56.9 & 48.1 & – \\
BEVDepth~\cite{li2022bevdepth} & AAAI'23 & C & - & - & 60.0 & 50.3 &  – \\
BEVFormer v2~\cite{yang2022bevformer} & CVPR'23 & C & - & - & 63.4 & 55.6 & – \\
\midrule
SECOND~\cite{yan2018second}       &Sensors'18 & L & 63.0 & 52.6 & 63.3 & 52.8 & 53.2 \\
PointPillars~\cite{lang2019pointpillars} &CVPR'19 & L & 61.3 & 52.3  & –    & –   & 28.1 \\
CenterPoint~\cite{yin2021cvpr}  & CVPR'21 & L & 66.8 & 59.6  & 67.3 & 60.3 & 62.7 \\
\midrule
PointPainting~\cite{vora2020pointpainting} & CVPR'20 & C+L & 69.6 & 65.8 & - & - & 151.8  \\
PointAugmenting~\cite{wang2021pointaugmenting} & CVPR'21 & C+L & - & - & 71.0$^\dag$ & 66.8$^\dag$ & 188.4 \\
MVP~\cite{yin2021multimodal}  & NeurIPS'21 & C+L & 70.0 & 66.1 & 70.5 & 66.4 & 148.1 \\
FusionPainting~\cite{xu2021fusionpainting} &ITSC'21 & C+L & 70.7 & 66.5 & 71.6 & 68.1 & -\\
FUTR3D~\cite{chen2022futr3d} & ArXiv'22 & C+L & 68.3 & 64.5 & - & - & 302.6 \\
AutoAlign~\cite{chen2022autoalign} & IJCAI'22 & C+L & 71.1 & 66.6 & - & - & -\\
TransFusion~\cite{bai2022transfusion} & CVPR'22 & C+L & 71.3 & 67.5 & 71.6 & 68.9 & 164.6 \\
AutoAlignV2~\cite{chen2022autoalignv2} & ECCV'22 & C+L & 71.2 & 67.1 & 72.4& 68.4&207.0 \\
UVTR~\cite{li2022unifying} & NeurIPS'22 & C+L & 70.2 & 65.4 & 71.1 & 67.1  & - \\
BEVFusion~(PKU)~\cite{liang2022bevfusion} & NeurIPS'22 & C+L & 71.0 & 67.9 & 71.8 & 69.2 & 1231.0 \\
DeepInteraction~\cite{yang2022deepinteraction} & NeurIPS'22 & C+L & 72.6 & 69.9 & 73.4 & \textbf{70.8} & 541.1 \\
BEVFusion~(MIT)~\cite{liu2022bevfusion} & ICRA'23 & C+L & 71.4 & 68.5 & 72.9 & 70.2 & 130.5 \\
UniTR~(Ours) & ICCV'23 & C+L & \textbf{73.1} & \textbf{70.0} & \textbf{74.1} & 70.5 & \textbf{88.7 (50.2$^\ddag$)} \\
UniTR w/ LSS~(Ours) & ICCV'23 & C+L & \textbf{73.3} & \textbf{70.5} & \textbf{74.5} & \textbf{70.9} & \textbf{107.5 (69.1$^\ddag$)} \\
\bottomrule
\end{tabular}}
\end{center}
\vspace{-8pt}
\caption{Performance of 3D detection on nuScenes (val and test) dataset~\cite{caesar2020nuscenes}. Notion of modality: Camera (C), LiDAR (L). $\dag$: with test-time augmentation. $\ddag$: deployed by TensorRT). We highlight the top-2 entries with \textbf{bold} font in each column.}
\label{tab:nus_det}
\vspace{-12pt}
\end{table*}
\subsection{Perception Task Setup}\label{sec:multi-task}
UniTR can be adapted to most 3D perception tasks. To demonstrate its versatility, we evaluate it on two important tasks: 3D object detection and BEV map segmentation.

\noindent \textbf{Detection.} Without loss of generality, we follow the same detection framework adopted in BEVFusion~\cite{liu2022bevfusion}. Notably, we only switch its multiple modality-specific encoders to ours and remove the redundant late fusion module while all other settings remain unchanged. To fully exploit geometrically enhanced image features, we also present an augmented variant  further strengthened by an additional LSS-based BEV fusion step~\cite{liu2022bevfusion,liang2022bevfusion}. More implementation details are described in previous 3D detection papers~\cite{liu2022bevfusion,bai2022transfusion}. 

\noindent \textbf{Segmentation.} We adopt the same framework as our 3D object detection task, with the exception of segmentation head and evaluation protocol used in BEVFusion~\cite{liu2022bevfusion}.

\section{Experiments}
\begin{table*}[h]
\begin{center}
\resizebox{0.85\linewidth}{!}{
\begin{tabular}{l|c|cccccc|>{\columncolor{mygray}}c}
\toprule
 Methods & Modality & Drivable & Ped. Cross. & Walkway & Stop Line & Carpark & Divider & Mean IoU \\
\midrule
OFT~\cite{roddick2018orthographic} & C & 74.0 & 35.3 & 45.9 & 27.5 & 35.9 & 33.9 & 42.1 \\
LSS~\cite{philion2020lift} & C & 75.4 & 38.8 & 46.3 & 30.3 & 39.1 & 36.5 & 44.4 \\
CVT~\cite{zhou2022cross} & C & 74.3 & 36.8 & 39.9 & 25.8 & 35.0 & 29.4 & 40.2 \\
M$^2$BEV~\cite{xie2022m} & C & 77.2 & – & – & – & – & 40.5 & – \\
BEVFusion~\cite{liu2022bevfusion} & C & 81.7 & 54.8 & 58.4 & 47.4 & 50.7 & 46.4 & 56.6 \\
\midrule
PointPillars~\cite{lang2019pointpillars} & L & 72.0 & 43.1 & 53.1 & 29.7 & 27.7 & 37.5 & 43.8 \\
CenterPoint~\cite{yin2021cvpr}  & L & 75.6 & 48.4 & 57.5 & 36.5 & 31.7 & 41.9 & 48.6 \\
UniTR (Ours) & L & \textbf{89.6} & \textbf{71.4} & \textbf{75.8} & \textbf{64.7} & \textbf{64.4} & \textbf{61.5}  & \textbf{71.2} \\
\midrule
PointPainting~\cite{vora2020pointpainting} & C+L & 75.9 & 48.5 & 57.1 & 36.9 & 34.5 & 41.9 & 49.1 \\
MVP~\cite{yin2021multimodal} & C+L & 76.1 & 48.7 & 57.0 & 36.9 & 33.0 & 42.2 & 49.0 \\
BEVFusion~\cite{liu2022bevfusion} & C+L & 85.5 & 60.5 & 67.6 & 52.0 & 57.0 & 53.7 & 62.7 \\
UniTR~(Ours) & C+L & \textbf{90.4} & \textbf{73.1} & \textbf{78.2} & \textbf{66.6} & \textbf{67.3} & \textbf{63.8}  & \textbf{73.2} \\
UniTR w/ LSS~(Ours) & C+L & \textbf{90.5} & \textbf{73.8} & \textbf{79.1} & \textbf{68.0} & \textbf{72.7} & \textbf{64.0}  & \textbf{74.7} \\
\bottomrule
\end{tabular}}
\end{center}
\vspace{-4pt}
\caption{UniTR outperforms the state-of-the-art multi-sensor fusion methods on BEV map segmentation on nuScenes (val), which demonstrates the effectiveness of our unified backbone for semantic-oriented 3D perception tasks. All baselines are reported in BEVFusion~\cite{liu2022bevfusion}. }
\label{tab:nus_seg}
\vspace{-10pt}
\end{table*}

\begin{table}[t]
\centering
\resizebox{1.0\linewidth}{!}{
\begin{tabular}{cccccc}
\toprule
Modality & 3D Space~(L) & 2D Space~(C)  & BEVLSS & NDS & mAP \\
\midrule
L & & & & 70.5 & 65.9 \\
C+L &\checkmark & & & 72.0 & 68.5 \\
C+L && \checkmark & & 72.5 & 69.0 \\
C+L &\checkmark & \checkmark & & 73.1& 70.0 \\
C+L &\checkmark & \checkmark & \checkmark & 73.3 & 70.5 \\
\toprule
\end{tabular}}
\vspace{-4pt}
\caption{Effect of camera image space fusion, 3D lidar geometric space fusion and BEV unifier on nuScenes (val). 1$^{st}$ row is the lidar-only variant of our model. Camera~(C), LiDAR~(L).}
\label{tab:fuse}
\vspace{-6pt}
\end{table}
\begin{table}[h]
\centering
\resizebox{0.9\linewidth}{!}{
\begin{tabular}{cccccc}
\toprule
Modality & Serial & Parallel  & Latency(ms) & NDS & mAP \\
\midrule
C        &           &           &  28          & 36.2 & 31.4 \\
L         &           &           &  26          & 70.5 & 65.9 \\ 
C+L  &\checkmark &           &  51          & 72.2 & 68.5 \\
C+L  &           &\checkmark &  33          & 72.4 & 69.0 \\
\bottomrule
\end{tabular}}
\vspace{-0pt}
\caption{Effect of parallel intra-modal transformer block.}
\label{tab:singlemodal}
\vspace{-16pt}
\end{table}
In this section, we conduct experiments on 3D object detection and BEV map segmentation tasks using nuScenes dataset~\cite{caesar2020nuscenes}, covering both geometric- and semantic-oriented 3D perceptions. Actually, owing to the inclusive 2D perspective and 3D geometric spaces, our unified backbone also offers versatility and seamless extensibility to accommodate other sensor types, \textit{e.g.}, radars and ultrasonic, as well as various other 3D perception tasks, \emph{e.g.}, 3D object tracking~\cite{yin2021cvpr} and motion prediction~\cite{ettinger2021large}.
\subsection{Implementation Details}
\noindent \textbf{Backbone.} Our UniTR starts with two modality-specific tokenizers, a DVFE layer~\cite{zhou2020end} for point clouds voxelization with grid size (0.3m, 0.3m, 8m) of detection and (0.4m, 0.4m, 8m) of segmentation, and an image patch tokenizer~\cite{dosovitskiy2020image} for 8$\times$ downsampling camera images to 32$\times$88. Then one weight-sharing intra-modal UniTR block is followed for processing modal-wise representation learning in parallel. Finally, three inter-modal UniTR blocks cross the boundaries of different modalities and provide connection among them by alternating between 2D and 3D partitioning configurations consecutively. The block configuration adopted in this paper is \{intra, inter$_{2D}$, inter$_{2D}$, inter$_{3D}$\}. More elaborated details are in Appendix. 

\noindent \textbf{Dataset.} We conduct our experiments on nuScenes~\cite{caesar2020nuscenes}, a challenging large-scale outdoor benchmark that provides diverse annotations for various tasks, (\emph{e.g.}, 3D object detection~\cite{yin2021cvpr} and BEV map segmentation~\cite{liu2022bevfusion,li2022hdmapnet}). It contains 40,157 annotated samples, each with six monocular camera images covering a 360-degree FoV and a 32-beam LiDAR.

\noindent \textbf{Training.} Unlike the dominant two-step training strategies~\cite{liu2022bevfusion,liang2022bevfusion,bai2022transfusion} that contain separate single-modal pre-training and joint multi-modal post-training, our UniTR is trained by a simpler one-step training scheme in an end-to-end manner with aligned multi-modal data augmentations. All the experiments are trained by AdamW optimizer~\cite{loshchilov2018decoupled} on 8 A100 GPUs. See appendix for more details. 
\subsection{3D Object Detection}
\noindent \textbf{Setting.} We benchmark UniTR on nuScenes~\cite{caesar2020nuscenes} dataset, which contains 1.4 million annotated 3D bounding boxes for 10 classes. Without loss of generality, we follow the framework of BEVFusion~\cite{liu2022bevfusion} and place our UniTR before its final multi-modal BEV encoder. The nuScenes detection score (NDS) and mean average precision (mAP) are reported. We also measure the latency of all open-source methods on the same workstation with an A100 GPU. Notably, we only use a single model without any test-time augmentation for both validation and test results.

\noindent \textbf{Results.} As shown in Table \ref{tab:nus_det}, our UniTR outperforms all previous LiDAR-camera fusion methods with much lower inference latency~(88.7ms). It achieves state-of-the-art performance, 73.1 and 74.1 in \textit{val} and \textit{test} NDS, surpassing BEVFusion~\cite{liu2022bevfusion} by +1.7 and +1.2 separately. After incorporating an additional LSS-based BEV fusion step, our UniTR achieves an enhanced performance, reaching 73.3 and 74.5 in \textit{val} and \textit{test} NDS, respectively. Notably, our model is built upon DSVT~\cite{wang2023dsvt} and inherits its deployment-friendly characteristic, making it easily accelerated by well-optimized deployment tools (\textit{i.e.}, TensorRT) to lower inference latency~(50.2ms). To the best of our knowledge, our method is the first to achieve state-of-the-art performance with close-to-real-time running speed (about 20 \textit{Hz}).
\subsection{BEV Map Segmentation} 
\noindent \textbf{Setting.} To further demonstrate the generalizability of our approach, we follow BEVFusion~\cite{liu2022bevfusion} and evaluate our model on BEV Map Segmentation task. This segmentation variant is adapted from our multi-modal detection approach, and replaced by the perception head and evaluation protocol used in BEVFusion. The per-class and averaged mean IoU scores are reported for our evaluation metric. 

\noindent \textbf{Results.} We report the comparison results with state-of-the-art methods in Table \ref{tab:nus_seg}. Our unitr and enhanced variant achieve 73.2 and 74.7 mIoU on nuScenes validation set respectively, which outperforms previous best result~\cite{liu2022bevfusion} by +10.5 and +12.0, demonstrating its effectiveness of semantic-oriented 3D perception tasks.
\subsection{Ablation Studies and Discussions} We conduct comprehensive ablation studies on the validation set of nuScenes 3D object detection to analyze individual components of our proposed method. Notably, all the experiments are performed on our BEV enhanced variant.

\noindent \textbf{Effect of 2D \& 3D fusion.} We first ablate the effects of our 2D \& 3D cross-modal representation learning blocks in Table \ref{tab:fuse}. The base competitor (1$^{st}$ row) is our lidar-only variant that abandons image information. To make a fair comparison, we only switch  the fusion algorithm while all other settings remain unchanged (such as the number of transformer blocks). See appendix for more details. As evidenced in the 1$^{st}$, 2$^{nd}$ and 3$^{rd}$ rows, processing cross-modal interaction by considering the perspective 2D image space and 3D sparse space separately brings considerable performance gains of our model, \emph{i.e.}, 70.5 $\rightarrow$ 72.5, 70.5 $\rightarrow$ 72.0 on NDS. Combining both of them (4$^{th}$ rows), our model can achieve a significant improvement, \emph{i.e.}, 70.5 $\rightarrow$ 73.1, 65.9 $\rightarrow$ 70.0 on NDS and mAP. That verifies our motivation that the semantic-rich 2D perspective view and geometric-sensitive 3D view are two complementary spaces for boosting 3D perception performance. 

\begin{table}[t]
    \centering
    \resizebox{\linewidth}{!}{
    \begin{tabular}{c|cc|cc|cc|cc}
    \toprule
         & \multicolumn{2}{|c|}{Clean} & \multicolumn{2}{c|}{Missing F} & \multicolumn{2}{c|}{Preserve F} & \multicolumn{2}{c}{Stuck} \\
         Approach & mAP & NDS &mAP & NDS & mAP & NDS &mAP& NDS  \\
         \midrule
         DETR3D*~\cite{wang2022detr3d}& 34.9 & 43.4 & 25.8& 39.2& 3.3&20.5&17.3&32.3\\
         PointAugmenting~\cite{wang2021pointaugmenting} & 46.9 & 55.6&42.4&53.0&31.6&46.5&42.1&52.8 \\
         MVX-Net~\cite{sindagi2019mvx} & 61.0&66.1&47.8&59.4&17.5&41.7&48.3&58.8\\
         TransFusion~\cite{bai2022transfusion} &66.9&70.9&65.3&70.1&64.4&69.3&65.9&70.2 \\
         BEVFusion~\cite{liang2022bevfusion} & 67.9& 71.0&65.9&70.7&65.1&69.9&66.2&70.3 \\
         \midrule
         UniTR~(Ours) & 70.5&73.3&68.5&72.4&66.5&71.2&68.1&71.8\\
         \bottomrule
    \end{tabular}}
    \vspace{-4pt}
    \caption{Results on robustness setting of camera failure cases. F denotes the front camera, and * means camera-only inputs. All the experiments are on nuScenes validation set.}
    \label{tab:camerarobust}
    \vspace{-8pt}
\end{table}
\begin{table}[t]
    \centering
    \resizebox{0.8\linewidth}{!}{
    \begin{tabular}{ccccc}
    \toprule
    Aug & Metrics & LiDAR & BEVFusion\cite{liang2022bevfusion} & Ours \\
    \midrule
               &   mAP & 31.3 & 40.2 (+8.9) & 38.3(+7.0) \\
               &   NDS & 50.7 & 54.3 (+3.6)) & 55.6(+4.9) \\
    \midrule
    \checkmark &   mAP & - & 54.0(+22.7) & 60.2(+28.9) \\
    \checkmark &   NDS & - & 61.6(+10.9) & 66.0(+15.3) \\
        \bottomrule
    \end{tabular}}
    \caption{Results on robustness setting of object failure cases. We refer readers to \cite{liang2022bevfusion} for more details.}
    \label{tab:obf_ab}
    \vspace{-16pt}
\end{table}
\noindent \textbf{Effect of parallel intra-modal transformer block.} Table \ref{tab:singlemodal} ablates the effectiveness of our parallel intra-modal transformer block. The 1$^{st}$ and 2$^{nd}$ rows are the pure-image and pure-lidar baselines of our method. To better evaluate the effectiveness of this weight-sharing manner, we remove the powerful 2D\&3D fusion algorithm except for the final late fusion module and restrict backbone to only carry out intra-modal representation learning within each modality. By comparing the 3$^{rd}$ and 4$^{th}$ rows, we find that weight-sharing backbone can achieve slightly better performance than previous sequential manner with much lower inference latency, \emph{i.e}, 51ms $\rightarrow$ 33ms.  We argue that the unified encoder is naturally suitable for aligning representations of different modalities, especially for the closely related and complementary image-lidar pairs, which encourages our model to overcome the modality gap and learn generic representations for 3D outdoor perception more easily. 

As for the faster running speed, attention computation is very fast, so involving more tokens in this operation doesn't significantly slow down the inference, \emph{e.g.}, lidar-only~(26ms) vs parallel image-lidar~(33ms). However, calling attention frequently will introduce some additional overheads, (\textit{e.g.}, memory access), and increase the latency.

\noindent \textbf{Effect of different block configurations.} All the blocks in UniTR are independent and can be flexibly configured. Table \ref{tab:block_config} presents the results for various block setups. The "Inter $\rightarrow$ Intra" configuration exhibits inferior performance, possibly due to the challenges encountered when fusing modalities for shallow features. Similarly, the "3D $\rightarrow$ 2D" configuration shows slightly lower performance.  We argue that it is more advantageous for the last block to perform fusion in the 3D space, as it aligns with the necessity of converting features into 3D for perception.
\subsection{Robustness Against Sensor Failure}
We follow the same evaluation protocols adopted in BEVFusion~\cite{liang2022bevfusion} to demonstrate the robustness of our UniTR for lidar and camera malfunctioning. We refer readers to \cite{yu2022benchmarking,liang2022bevfusion} for more implementation details. All the experiments are conducted on nuScenes validation set. 

As shown in Table \ref{tab:camerarobust} and \ref{tab:obf_ab}, our method outperforms all the single- and multi-modal methods regardless of any LiDAR or camera malfunction scenarios, which demonstrates the robustness of UniTR against sensors failure cases. To further illustrate the motivation behind unification, the combination of cameras with low-beam LiDAR has been conducted. We followed the same setting adopted in BEVFusion~\cite{liu2022bevfusion}. As depicted in Table \ref{tab:low_beam_ab}, our unified modeling can better leverage the complementary information among different sensors, particularly benefiting the sparser LiDAR and leading to enhanced performance robustness.

\begin{table}[t]
   \scriptsize
   \scalebox{0.98}{
  	\begin{minipage}[b]{0.55\linewidth}
  		\centering
  		\makeatletter\def\@captype{table}
  		\setlength{\tabcolsep}{1mm}{
    	\begin{tabular}{lcccc}
        	\toprule
        	Modality & Camera & Lidar & Serial & Parallel \\
        	\hline
        	\noalign{\smallskip}
        	C+L(1-beam)   & 36.2 & 42.0 & 57.6 & \textbf{59.5} \\
        	C+L(4-beam)   & 36.2 & 62.1 & 67.3 & \textbf{68.5} \\
        	C+L(16-beam)  & 36.2 & 69.1 & 71.6 & \textbf{72.2} \\
                C+L(32-beam)  & 36.2 & 70.5 & 73.2 & \textbf{73.3} \\
        	\bottomrule
    	\end{tabular}}
  		\caption{Ablation of low beam setting with NDS evaluation metric. }
  		\label{tab:low_beam_ab}
  		\vspace{-8pt}
    \end{minipage}
    \hspace{8pt}
  	\begin{minipage}[b]{0.4\linewidth}
  		\centering
  		\makeatletter\def\@captype{table}
  		\setlength{\tabcolsep}{1mm}{
    	\begin{tabular}{ccc}
        	\toprule
        	Block config & NDS & mAP\\
        	\hline
        	\noalign{\smallskip}
                3D $\rightarrow$ 2D  & 73.0 & 70.0 \\
                2D $\rightarrow$ 3D  & \textbf{73.3} & \textbf{70.5} \\
        	  Inter $\rightarrow$ Intra  & 72.9 & 69.8\\
                Intra $\rightarrow$ Inter  & \textbf{73.3} & \textbf{70.5}\\
        	\bottomrule
    	\end{tabular}}
  		\caption{Results of different block configurations. }
  		\label{tab:block_config}
  		\vspace{-8pt}
    \end{minipage}}
\end{table}
\begin{table}[t]
\begin{center}
    \centering
        \resizebox{0.8\linewidth}{!}{
        \begin{tabular}{cccc}
        \toprule
        Models & Latency~(ms) & NDS & mAP \\
        \midrule
        PointPainting~\cite{vora2020pointpainting} & 151.8 & 69.6 & 65.8\\
        TransFusion~\cite{bai2022transfusion} & 164.6 & 71.3 & 67.5\\
        BEVFusion~(PKU)~\cite{liang2022bevfusion} & 1231.0 & 71.0 & 67.9\\
        BEVFusion~(MIT)~\cite{liu2022bevfusion} & 130.5 & 71.4 & 68.5\\
        UniTR & \textbf{88.7} &  \textbf{73.1} & \textbf{70.0}\\
        UniTR~(TensorRT) &\textbf{50.2} & \textbf{73.1} & \textbf{70.0}\\
        \bottomrule
        \end{tabular}}
\end{center}
\vspace{-8pt}
\caption{The latency and performance on nuScenes val set.}
\label{table:acc_speed}
\vspace{-16pt}
\end{table}
\subsection{Inference Speed}
We present a comparison with other state-of-the-art methods on both inference latency and performance accuracy in Table \ref{table:acc_speed}. Our model significantly outperforms BEVFusion~\cite{liu2022bevfusion} while achieving a lower latency, \emph{i.e.}, 88.7 ms vs 130.5 ms. Thanks to its deployment-friendly characteristic, our model can reach a close-to-real-time running speed~(about 20\textit{Hz}) after being deployed by TensorRT. All the experiments are evaluated on the same workstation.

\section{Conclusion}
In this paper, we propose a unified backbone that processes various modalities with a single model and shared parameters for outdoor 3D perception. With the specially designed modality-agnostic transformer blocks for intra- and inter-modal representation learning, our method can achieve state-of-the-art performance with remarkable gains of various 3D perception tasks on challenging nuScenes dataset. It is our belief that UniTR can provide a strong foundation for facilitating the development of more efficient and generic outdoor 3D perception systems. 

\noindent \textbf{Acknowledgments}. This work is supported by National Key R\&D Program of China (2022ZD0114900) and National Science Foundation of China (NSFC62276005). We gratefully acknowledge the support of Mindspore, CANN(Compute Architecture for Neural Networks) and Ascend AI Processor used for this research. 

{\small
\bibliographystyle{ieee_fullname}
\bibliography{egbib}
}

\newpage
\appendix

In our supplementary material, we offer in-depth insights into our network architecture, training methodologies, and ablation baselines, which can be found in Section \ref{sec:imple_app}. Moreover, we present an extensive examination of our robustness experiments in Section \ref{sec:robust}. Lastly, we address the limitations of UniTR in Section \ref{sec:limit}.
\section{Implementation Details} \label{sec:imple_app}
\subsection{Network Architecture}
\noindent \textbf{Tokenizers}. In outdoor perception scenarios, we consider two input modalities: multi-view camera images and LiDAR point clouds. For image inputs, we first take the raw images $X^I \in \mathbb{R}^{6 \times 256 \times 704 \times 3}$ captured by six cameras and split them into non-overlapping patches using a patch-splitting module, similar to ViT~\cite{dosovitskiy2021an}. Each patch serves as a "token" with its feature being a concatenation of raw pixel RGB values. In our implementation, we employ an $8 \times 8$ patch size, resulting in a feature dimension of $8 \times 8 \times 3 = 192$. A linear embedding layer is then applied to these features, projecting them to an arbitrary dimension denoted as $C$. The output of the image tokenizer is $\mathcal{T}^I \in \mathbb{R}^{M \times C}$, where $M$ represents the token number.

Regarding LiDAR point clouds $X^P$, we utilize the standard dynamic voxel feature encoding tokenizer~\cite{zhou2020end} as implemented by OpenPCDet~\cite{openpcdet2020}. We use a grid size of $(0.3m, 0.3m, 8.0m)$ for detection and $(0.4m, 0.4m, 8.0m)$ for segmentation to generate LiDAR voxels, $\mathcal{T}^{P} \in \mathbb{R}^{N \times C}$. By employing these two tokenizers, the multi-modal inputs can be converted to $\mathcal{T} \in \mathbb{R}^{(M+N) \times C}$, which includes $N$ point cloud tokens and $M$ image tokens for subsequent intra-modal transformer blocks.

\noindent \textbf{Multi-modal Backbone}. Our UniTR features a single-stride, pillar-based multi-modal backbone that starts with one weight-sharing intra-modal transformer block for parallel processing of modal-wise representation learning. Subsequently, three inter-modal transformer blocks bridge different modalities and establish connections among them by alternating between 2D and 3D partitioning configurations. The block configuration adopted in this paper is \{intra, inter$_{2D}$, inter$_{2D}$, inter$_{3D}$\}. The window sizes for both $L^P \times W^P \times H^P$ and $L^I \times W^I \times 1$ are (30, 30, 1), and the maximum number of tokens assigned to each set ($\tau$) is set to 90 for all modalities. All attention modules are equipped with 8 heads, 128 input channels, and 256 hidden channels. For the inter-modal block (3D), the pseudo grid points size, $L^S \times W^S \times H^S$, is set to $360 \times 360 \times 20$.

\subsection{Ablation Baselines}
\noindent \textbf{Effect of 2D \& 3D fusion.} The base competitor is the lidar-only variant of our model with four intra-modal blocks~\cite{wang2023dsvt} and transfusion head~\cite{bai2022transfusion}. For a fair comparison, we only switch the fusion algorithm while keeping all other settings remain unchanged. The number of the intra- and inter-modal blocks is summarized in Table \ref{tab:fuse_setting}.

\begin{table}[h]
\centering
\resizebox{1.0\linewidth}{!}{
\begin{tabular}{ccccccc}
\toprule
Modality & Intra-B & Inter-B~(2D) & Inter-B~(3D) & BEVLSS & NDS & mAP \\
\midrule
L        &      4  &            0 &            0 &            &70.5 & 65.9 \\
C+L      &      3  &            0 &            1 &            &72.0 & 68.5 \\
C+L      &      3  &            1 &            0 &            &72.5 & 69.0 \\
C+L      &      2  &            1 &            1 &            &72.9 & 69.8 \\
C+L      &      1  &            2 &            1 &            &73.1 & 70.0 \\
C+L      &      1  &            2 &            1 &\checkmark  &73.3 & 70.5 \\
\toprule
\end{tabular}}
\vspace{-4pt}
\caption{The number of the intra- and inter-modal blocks on the ablation of 2D \& 3D fusion. Camera~(C), LiDAR~(L).}
\label{tab:fuse_setting}
\vspace{-4pt}
\end{table}

\noindent \textbf{Effect of parallel intra-modal transformer block.} The 1$^{st}$ and 2$^{nd}$ rows are the image-only and lidar-only baselines with four intra-modal blocks. To evaluate the effectiveness of our weight-sharing approach, we conducted experiments with both serial and parallel multi-modal variants, where only a BEV unifier was used without our proposed 2D\&3D fusion strategies. This allowed us to better isolate the impact of the weight-sharing approach on its own, separate from the strong fusion strategies. All latency measurements are taken on the same workstation with an A100 GPU. Note that the latency reported in Table {\color{red}4} of the main paper only refers to the transformer backbone, without including the modality-specific tokenizers and partitioning.
\subsection{Training Schemes} 
As stated in the main paper, previous approaches for multi-modal fusion usually involve two-step training strategies with separate single-modal pre-training and joint multi-modal post-training for fusion. In contrast, our UniTR can be directly trained with a one-step end-to-end training scheme, where the data augmentations of the image and lidar are aligned.  We follow the matching strategies of BEV and image space data augmentation used in BEVFusion~\cite{liu2022bevfusion}, \emph{e.g.}, random rotation, translation, and flip. To synchronize 2D-3D joint GT-AUG, we use the same implementation of cross-modal copy-paste proposed in \cite{chen2022autoalignv2} with a Mix-up Ratio $\alpha=0.7$ and add fade strategy at the last two epochs. Our UniTR backbone is pre-trained on both ImageNet~\cite{deng2009imagenet} and nuImage~\cite{caesar2020nuscenes} datasets.  We train all experiments using the AdamW optimizer~\cite{loshchilov2018decoupled} on 8 A100 GPUs with weight decay 0.03, a one-cycle learning rate policy~\cite{onecyc}, and a maximum learning rate of 3e-3. For 3D object detection, we used a batch size of 24 and trained for 10 epochs, while for BEV map segmentation, we used a batch size of 24 and trained for 20 epochs. All inference times were measured on the same workstation (single A100 GPU and AMD EPYC 7513 CPU).

\section{Robustness Against Sensor Failure} \label{sec:robust}
\subsection{LiDAR Malfunctions.} To assess the robustness of our framework, we conducted experiments on the nuScenes validation set under conditions where objects cannot reflect LiDAR points. Such situations may arise during rainy weather when the reflection rate of certain common objects falls below the LiDAR system's threshold. To simulate this scenario, we employed the same dropping strategy as \cite{liang2022bevfusion}: each frame has a 50\% chance of dropping objects, and each object has a 50\% chance of dropping the LiDAR points it contains.

As shown in the main paper, our UniTR outperforms both the LiDAR-only stream and previous fusion approaches, such as BEVFusion~\cite{liang2022bevfusion}, in terms of accuracy when detectors are evaluated without robustness augmentation. Furthermore, our method exhibits significant improvements when the detectors are fine-tuned on the robust augmented training set, outpacing BEVFusion by a substantial margin.

\subsection{Camera Malfunctions.} We performed additional experiments to evaluate the robustness of our UniTR backbone against camera malfunctions in three scenarios outlined in \cite{liang2022bevfusion}: i) missing front camera, ii) missing all cameras except the front, and iii) 50\% of camera frames stuck. As demonstrated in the main paper, UniTR surpasses other LiDAR-camera fusion methods and even camera-only methods in these scenarios, showcasing its resilience against camera malfunctions.

\section{Limitation} \label{sec:limit}
Despite its notable performance and processing speed in multi-modal 3D perception, our UniTR has certain limitations that warrant attention. First, as it inherits features from DSVT, UniTR is primarily a single-stride backbone designed for outdoor BEV perception, which constrains its adaptability to various other 3D perception tasks, such as indoor 3D perception. Second, UniTR is mainly focused on jointly processing different sensor types without considering the compatibility of transformation modalities for diverse scenarios. For instance, it does not accommodate switching to LiDAR-only, image-only, or image-LiDAR encoders during the inference stage. The design of a more modality-flexible backbone utilizing a Mixture-of-Experts approach remains an open challenge for the 3D perception community.

\end{document}